\documentclass[letterpaper, 10 pt, conference]{ieeeconf} 
\IEEEoverridecommandlockouts                              
\usepackage{graphics} 
\usepackage{epsfig} 
\usepackage{mathptmx} 
\usepackage{times} 
\usepackage{amsmath} 
\usepackage{amssymb}  
\usepackage{siunitx}
\usepackage[euler]{textgreek}
\usepackage{float}
\usepackage{svg}
\usepackage{cite}

\begin{document}
    \title{\Large \bf
    Thermodynamically-informed Air-based Soft Heat Engine Design
    }
  \author{
      Charles Xiao$^{1*}$, Luke F. Gockowski$^1$, Bolin Liao$^1$, Megan T. Valentine$^1$, and Elliot W. Hawkes$^1$ 
      \thanks{This work was supported in part by NSF Grant 1935327.}
  \thanks{$^1$Department of Mechanical Engineering, University of California, Santa Barbara, CA 93106.}
   \thanks{$^*$ Corresponding author. Email: charles\_xiao@ucsb.edu}
\thanks{\textcopyright 2021 IEEE. Personal use of this material is permitted.  Permission from IEEE must be obtained for all other uses, in any current or future media, including reprinting/republishing this material for advertising or promotional purposes, creating new collective works, for resale or redistribution to servers or lists, or reuse of any copyrighted component of this work in other works}
}

\maketitle

\begin{abstract}

Soft heat engines are poised to play a vital role in future soft robots due to their easy integration into soft structures and low-voltage power requirements. 
Recent works have demonstrated soft heat engines relying on liquid-to-gas phase change materials.
However, despite the fact that many soft robots have air as a primary component, soft air cycles are not a focus of the field.
In this paper, we develop theory for air-based soft heat engines design and efficiency, and demonstrate experimentally that efficiency can be improved through careful cycle design. We compare a simple constant-load cycle to a designed decreasing-load cycle, inspired by the Otto cycle. While both efficiencies are relatively low, the Otto-like cycle improves efficiency by a factor of 11.3, demonstrating the promise of this approach. 
Our results lay the foundation for the development of air-based soft heat engines as a new option for powering soft robots.
\end{abstract}

\section{Introduction}

Despite their long existence, heat engines--i.e., engines that convert heat into mechanical energy--remain central to power generation and sustainable energy harvesting, and enable most forms of modern transportation. Recently, interest in developing soft heat engines has risen--owing to their utility in soft robots. Typically comprising a heat source that fuels gas expansion or incites phase-change, these engines drive a volumetric expansion and thus deform the soft materials that encase them. Such engines are capable of large strokes, avoid the need for unwieldy pumps, and do not require high-voltage power sources.

Combustion-based heat engines have enabled several soft jumping robots. Using butane, methane, or fuel mixtures, researchers have demonstrated impressive jumps with relatively high efficiencies (i.e., 0.4-0.7\%) \cite{tolley2014untethered,shepherd2013using,bartlett20153d,loepfe2015untethered}. However, their complex construction (i.e., fuel tanks, valving, ignition systems, etc.), finite reactant volume, and explosive nature limit their use to applications where high power output is needed--i.e., jumping.

\begin{figure}
  \begin{center}
  \includegraphics[width=\columnwidth]{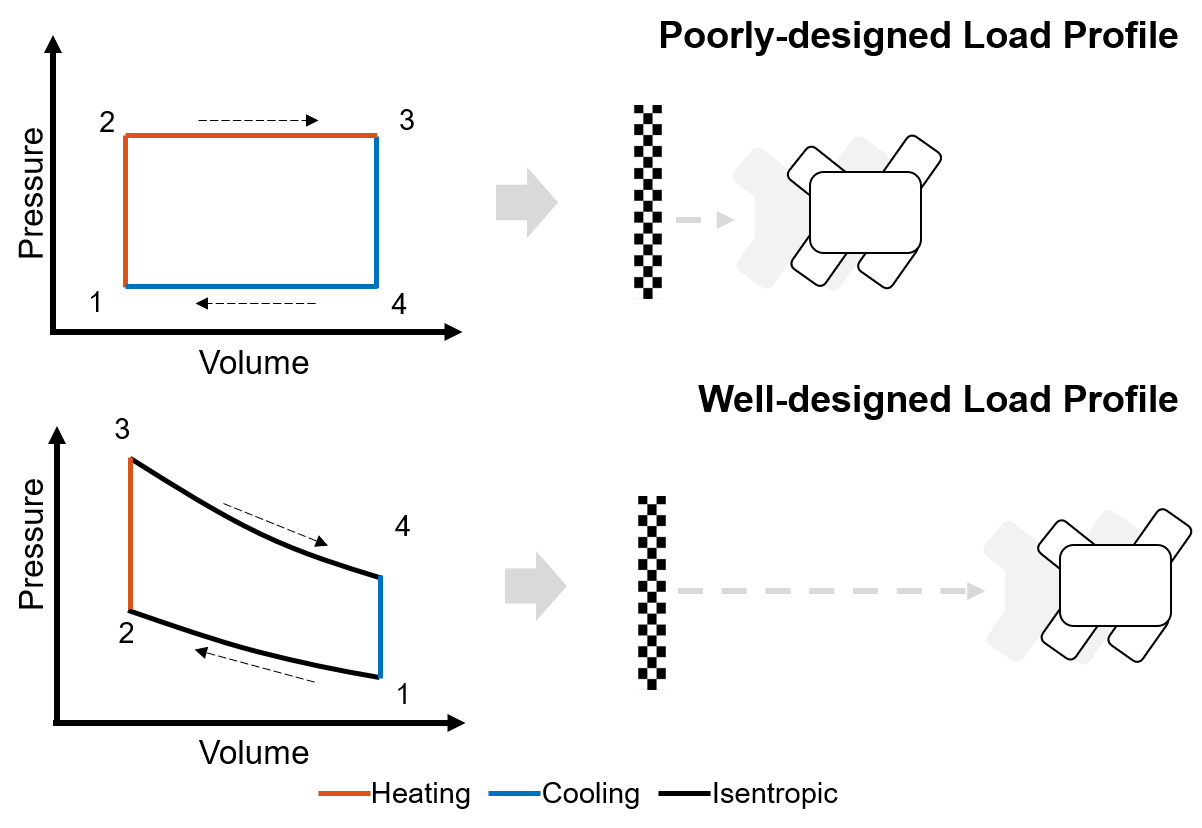}
  \caption{A well-designed load profile allows for more efficient thermodynamic cycles, resulting in longer soft robot operation.}
  \label{fig:graphAbstract}
  \end{center}
  \vspace{-3mm}
\end{figure}

Liquid-to-gas phase change heat engines are used widely in soft robots due to their simple implementation, ample force generation, and large strokes \cite{li2020soft,nishikawa2019design,kang2019programmable,chellattoan2020low,boyvat2019ultrastrong,narumi2020liquid,miriyev2017soft}. Such systems typically comprise a Joule heating element supplying heat to a low-boiling point fluid (e.g., Novec 7000 or Ethanol) contained in an elastic polymer pouch (e.g., EcoFlex 50, PDMS, thermoplastic polyurethane-coated fabrics). Impressive robots and actuators have been developed, but the limited selection of low-toxicity, low-boiling point fluids as well as chemically-resistant but highly-flexible pouch materials limit innovation. Further, reports of efficiency in this literature are scant, and when present consider only expansion \cite{boyvat2019ultrastrong} and not the efficiency of a full power cycle. 

Interestingly, even though many soft robots are comprised primarily of air (e.g., \cite{ohta2018design,qi2017design,usevitch2020untethered}), thermodynamic air cycles are nearly unexplored in the field. Air is non-toxic, abundant, and there exists a tome of rigorous theoretical analysis of high-efficiency air cycles in rigid systems (see \cite{moran2010fundamentals}). 
However, there are substantially different constraints on air-based soft heat engines than on rigid systems, such as low working temperatures and pressures, as well as the need for linear or non-rotary outputs.
Thus, there is a need to apply current thermodynamic theory to these soft air systems with their unique constraints.

Accordingly, here we study soft air-cycle heat engines.
Our work offers two main contributions.
First, we present theory on the efficiency of soft air systems given their constraints. 
Specifically, we show that the efficiency of a simple air cycle with constant-load expansion and compression can be improved by implementing decreasing-load stages, to resemble an Otto cycle. 
Second, we build and test an air heat engine, running it with both constant-load and decreasing-load cycles. We validate our theoretical results, and report practical concerns that further amplify the benefits of the decreasing-load cycle.
We note that the goal of this work is not to present a new actuator for immediate use in a robot. 
Rather, these contributions help advance our fundamental understanding of soft heat engines. Thus this work is foundational robotics research that lays the groundwork for future, more applied studies presenting ready-to-use actuators that can outperform the state-of-the-art.

\section{Modeling and Theoretical Results}

In this section,
we present thermodynamic analysis of air-based soft heat engines. We first discuss why a simple constant-load cycle, seen in some phase change actuators such as \cite{boyvat2019ultrastrong},
is inefficient when applied to air. 
We then give a detailed analysis of the constant-load cycle and our proposed Otto-like decreasing-load air cycle for soft robotics, showing the latter offers much higher efficiency.

\subsection{Load Design for air-based soft heat engines}

\begin{figure}
  \begin{center}
  \includegraphics[width=\columnwidth]{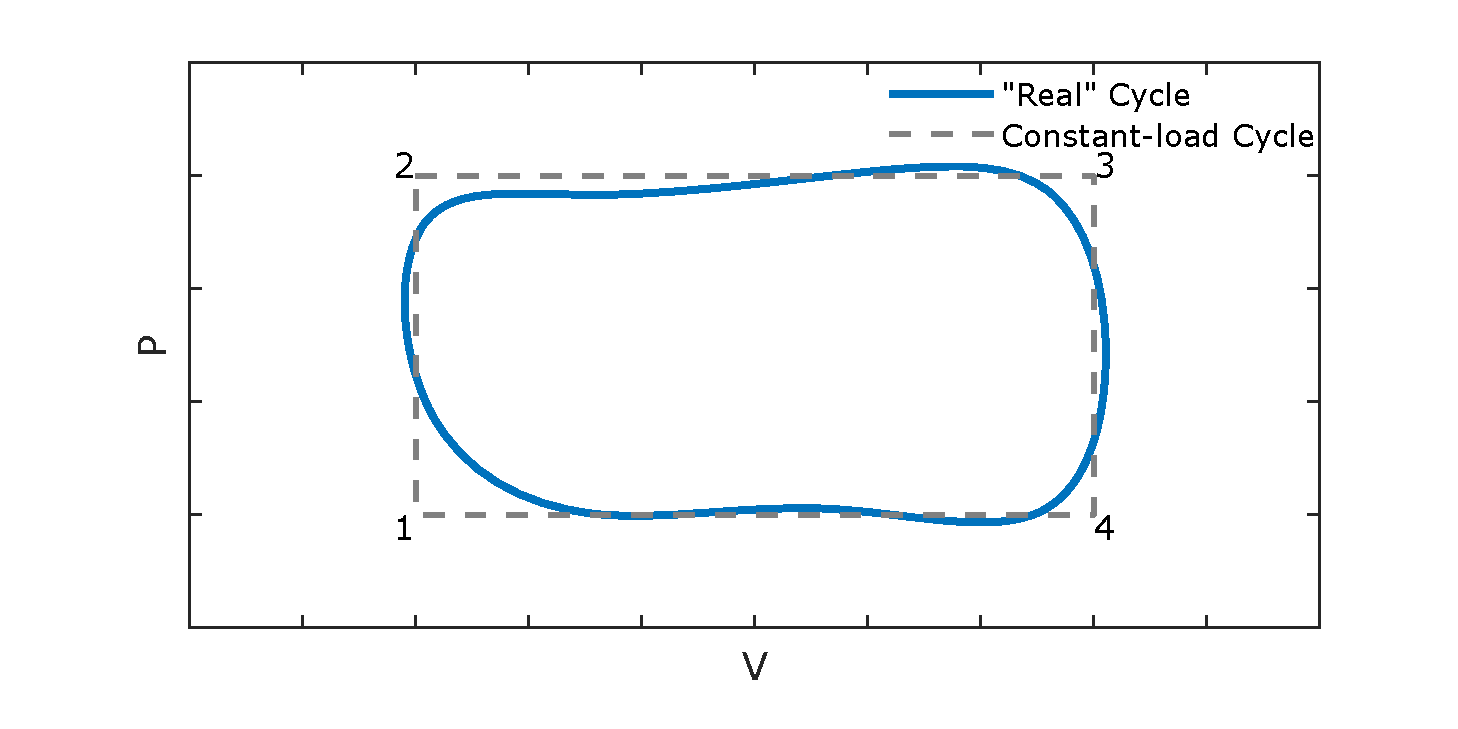}
  \caption{PV diagram showing a generalized soft heat engine cycle.  A constant-load cycle (dashed lined) is superimposed on top of it.}
  \label{fig:recCycle}
  \end{center}
\end{figure}

A simple, roughly constant-load heat engine cycle is illustrated in Fig. \ref{fig:recCycle}.
Specifically, the four stages are:
\begin{itemize}
    \item Isochoric (i.e., constant volume) heating from 1-2,
    \item Isobaric (i.e., constant load) expansion from 2-3,
    \item Isochoric cooling from 3-4, and
    \item Isobaric heat rejection from 4-1.
\end{itemize}

While these cycle designs are attractive due to their simplicity and ease of implementation, they are not the most efficient option. This is because the temperature of the air in the actuator must rise in order to maintain pressure during the isobaric expansion; non-isothermal heat transfer is inherently inefficient. Replacing this constant-load stage of the cycle with an isentropic expansion will increase the efficiency of the air cycle given the same maximum temperature. Similarly, replacing the isobaric compression with isentropic compression will further increase efficiency. This cycle resembles an Otto cycle \cite{moran2010fundamentals}.

To realize an approximation of an isentropic process in a soft heat engine, a rapid expansion or compression can be used (during which little heat is transferred). Critically, the load profile must be decreasing during the expansion, because a system's pressure (and temperature) drops as it expands.
Conversely, the load must increase during compression.

Thus, theoretically, efficiency can be improved by replacing the simple rectangular cycle that includes constant-load expansion and compression with an Otto-like cycle that includes a decreasing-load expansion and an increasing-load compression.

\subsection{Analysis of Air Cycles: Constant-load Versus Decreasing-load}

Here we provide a detailed analysis of the general insight provided in the previous section, first describing the efficiency of a constant-load air cycle, then that of a decreasing-load air cycle that resembles an Otto cycle, and finally comparing the two.

\subsubsection{Constant-load Cycle Analysis}
The dashed lines in Fig. \ref{fig:recCycle} shows a simply-designed cycle with a constant load profile (i.e., isobaric). 
In the most general form, the efficiency of such a cycle is
\[\eta=\frac{(P_2-P_1)(V_3-V_2)}{P_2(V_3-V_2)+U_3-U_1}\]
where $P$, $V$, and $U$ represents the pressure, volume, and internal energy, respectively. Qualitatively, the efficiency of such a system is lower than that of Carnot, because the working fluid experiences heat transfer at intermediate temperatures between the highest and the lowest temperatures.

For a system that runs on an ideal gas, the expression of efficiency in terms of volume and temperature is:
\[\eta=\frac{\left(\frac{T_3}{V_3}-\frac{T_1}{V_1}\right)(V_3-V_1)}{\frac{T_3}{V_3}(V_3-V_1)+c_v(T_3-T_1)}=\frac{\left(\frac{x}{r}-1\right)(r-1)}{\frac{x}{r}(r-1)+c_v(x-1)}\]
\[< 1-\frac{V_3T_1}{V_1T_3} =1-\frac{r}{x}\leq 1-\frac{T_1}{T_3}\]
where $c_v$, $T_1$, $T_3$, $r$, and $x$ are the dimensionless constant volume heat capacity, lowest temperature, highest temperature, expansion ratio ($r=V_3/V_1$), and temperature ratio ($x=T_3/T_1$), respectively. For air, $c_v\approx5/2$. The inequality shows that such a cycle has lower efficiency than Carnot and that the deviation from from the Carnot efficiency increases as the difference between $T_3$ and $T_1$ increases. 

\subsubsection{Decreasing-load Cycle Analysis}
Although the Carnot cycle is the most efficient thermodynamic cycle, it is challenging to realize. Consequently, real heat engines and their models, such as the Otto and Rankine cycles, deviate from the ideal by sacrificing efficiency for practicality (e.g., non-isothermal heat transfer) \cite{moran2010fundamentals}. For our paper, we chose an Otto-like cycle for simplicity. It requires one type of expansion and compression process rather than two (e.g., the Rankine cycle has isobaric and isentropic expansion and compression processes). The Otto engine is the idealization of the internal combustion engine. 

\begin{figure}
  \begin{center}
\includegraphics[width=\columnwidth]{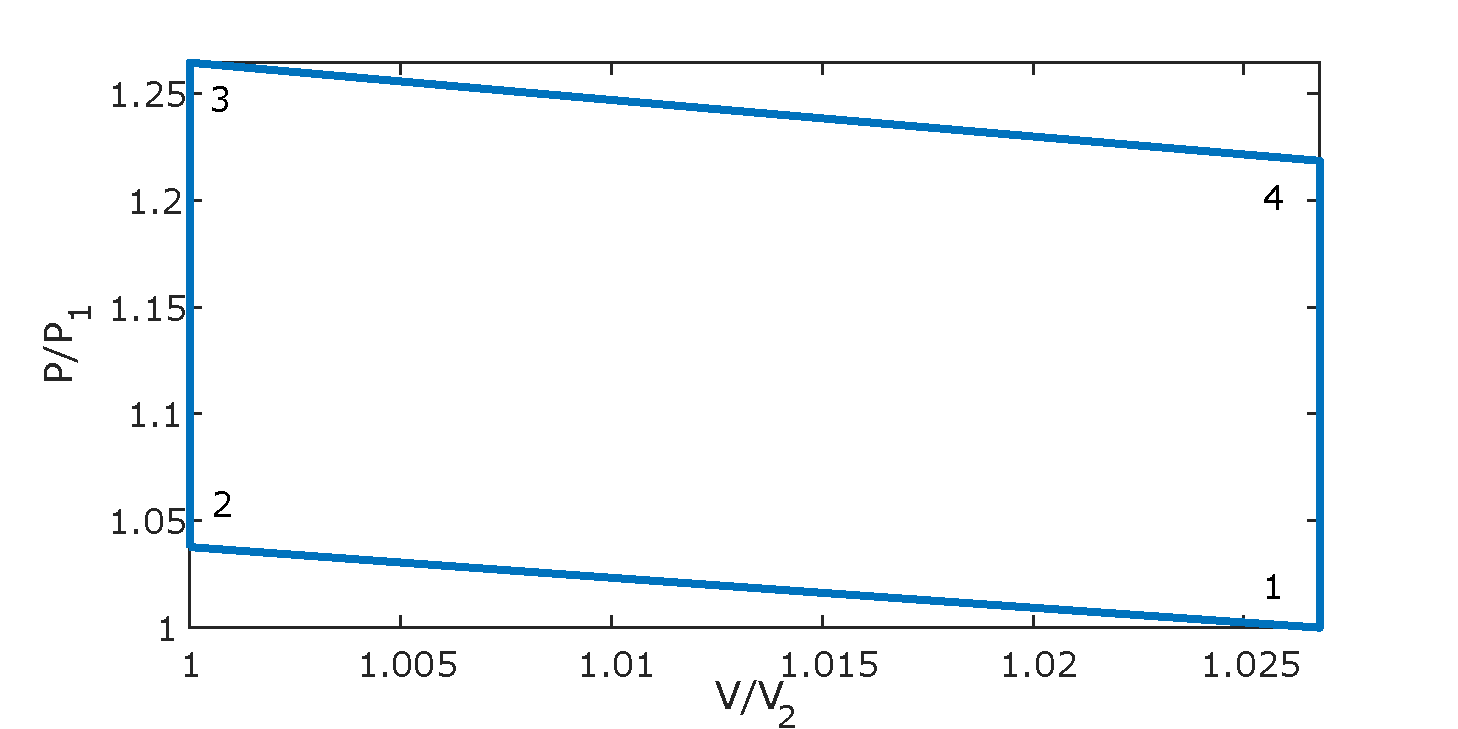}
  \caption{PV diagram of the Otto cycle drawn according to the experimental expansion ratio and target minimum and maximum pressures. At small expansions ratios and pressure differences, the shape resembles a parallelogram. }
  \label{fig:OttoSketch}
  \end{center}
\end{figure}

Fig. \ref{fig:OttoSketch} shows the Otto cycle, which we analyze as an approximation of our decreasing-load cycle; it consists of four parts:
\begin{itemize}
    \item Isentropic compression from 1-2
    \item Isochoric heating from 2-3
    \item Isentropic expansion from 3-4
    \item Isochoric heat rejection from 4-1.
\end{itemize}

Fig. \ref{fig:OttoSketch} was drawn according to the experimental expansion ratio and target minimum and maximum pressures. At low expansion ratios and pressure differences, the shape resembles a parallelogram, but in general the shape is more similar to the one shown in the bottom plot of Fig. \ref{fig:graphAbstract}. Note that the coordinates were chosen so that, like the constant-load cycle, the highest and lowest temperatures occur at points 3 and 1, respectively.

In the most general form, the efficiency of the Otto cycle is
\[\eta=\frac{W_{34}-W_{21}}{Q_{23}}=\frac{(U_3-U_4)-(U_2-U_1)}{U_3-U_2}=1-\frac{U_4-U_1}{U_3-U_2}\]
where $W$, $Q$, and $U$ represents the work done between two points, the heat addition between two points, and the internal energy at a given point. Qualitatively, the efficiency of the system improves as the compression ratio, $r=V_1/V_2=V_4/V_3$, increases, because the more the system expands the smaller the magnitude of $U_4$ and $U_1$ relatively to $U_3$ and $U_2$.
\par
The efficiency of an ideal gas based Otto cycle is
\[\eta=1-\frac{T_{1}}{T_{2}}=1-\frac{T_{4}}{T_{3}}=1-\left(\frac{V_2}{V_1}\right)^{\gamma-1}=1-\left(\frac{V_3}{V_4}\right)^{\gamma-1}\]
\[=1-\frac{1}{r^{\gamma-1}}\leq 1-\frac{T_1}{T_3}\]
where $T$, $V$, and $\gamma$ are the temperature, volume,  and specific heat ratio respectively.  For air, $\gamma\approx1.4$. For an ideal gas Otto cycle, increasing the expansion ratio explicitly increases the efficiency of the cycle \cite{moran2010fundamentals}. Overall, the efficiency is lower than Carnot, but Carnot efficiencies are approached when the difference between $T_3$ and $T_2$ is reduced.

\par

\subsubsection{Comparison of Efficiencies}
Typically, efficiencies are compared by matching the temperatures, but for the Otto and constant-load cycles, they also depend on volume. We compare the efficiencies by matching the expansion and temperature ratios. 

Before we compare the efficiencies, we note that for a given temperature, there are maximum expansion ratio constraints. For the constant-load and Otto cycles, the upper bounds of the expansion ratios are: 
\[r_{CL, max}<\frac{V_3}{V_1}=\frac{T_3}{T_1}\]
\[r_{O, max}<\frac{V_3}{V_1}=\left(\frac{T_3}{T_1}\right)^\frac{1}{\gamma-1}\]
\[r_{CL, max}<r_{O, max}\]

\begin{figure}
  \begin{center}
  \includegraphics[width=\columnwidth]{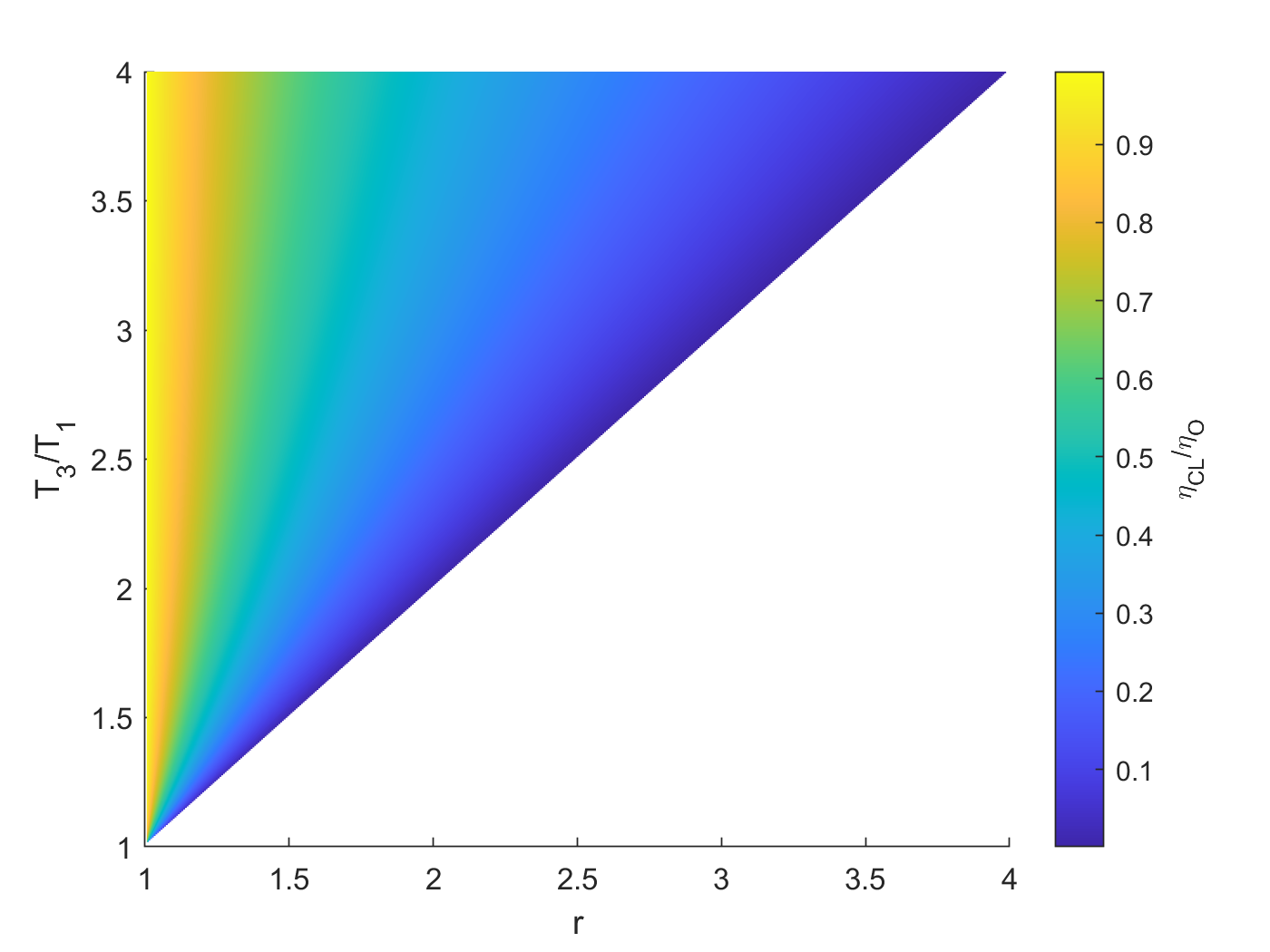}
  \caption{Heat map of the constant-load to Otto efficiency ratio for matched temperatures and expansion ratios. The blank regions are regions that are outside the maximum expansion ratio constraint for both the constant-load and Otto cycle. }
  
  \label{fig:ottoEff}
  \end{center}
\end{figure}

Fig. \ref{fig:ottoEff} plots the constant-load to Otto cycle efficiency ratio $\eta_{CL}/\eta_O$. From the plot it is evident that for a given maximum and minimum temperature, an Otto-like cycle is more efficient than a constant-load cycle for any valid constant-load expansion ratio. Moreover, the Otto cycle has a higher expansion ratio bound allowing it to achieve even higher efficiencies given the same temperature ratios.

\subsection{Key Theoretical Conclusions}
\begin{enumerate}
    \item Constant-load air processes are relatively inefficient.
    \item Isentropic processes can be used to improve upon constant-load cycle efficiency by converting heat to work and vice versa.
    \item Decreasing-load profiles are required for isentropic processes.

\end{enumerate}

\section{Experimental Methods}
\subsection{Testing Apparatus}

\begin{figure}
  \begin{center}
  \includegraphics[width=\columnwidth]{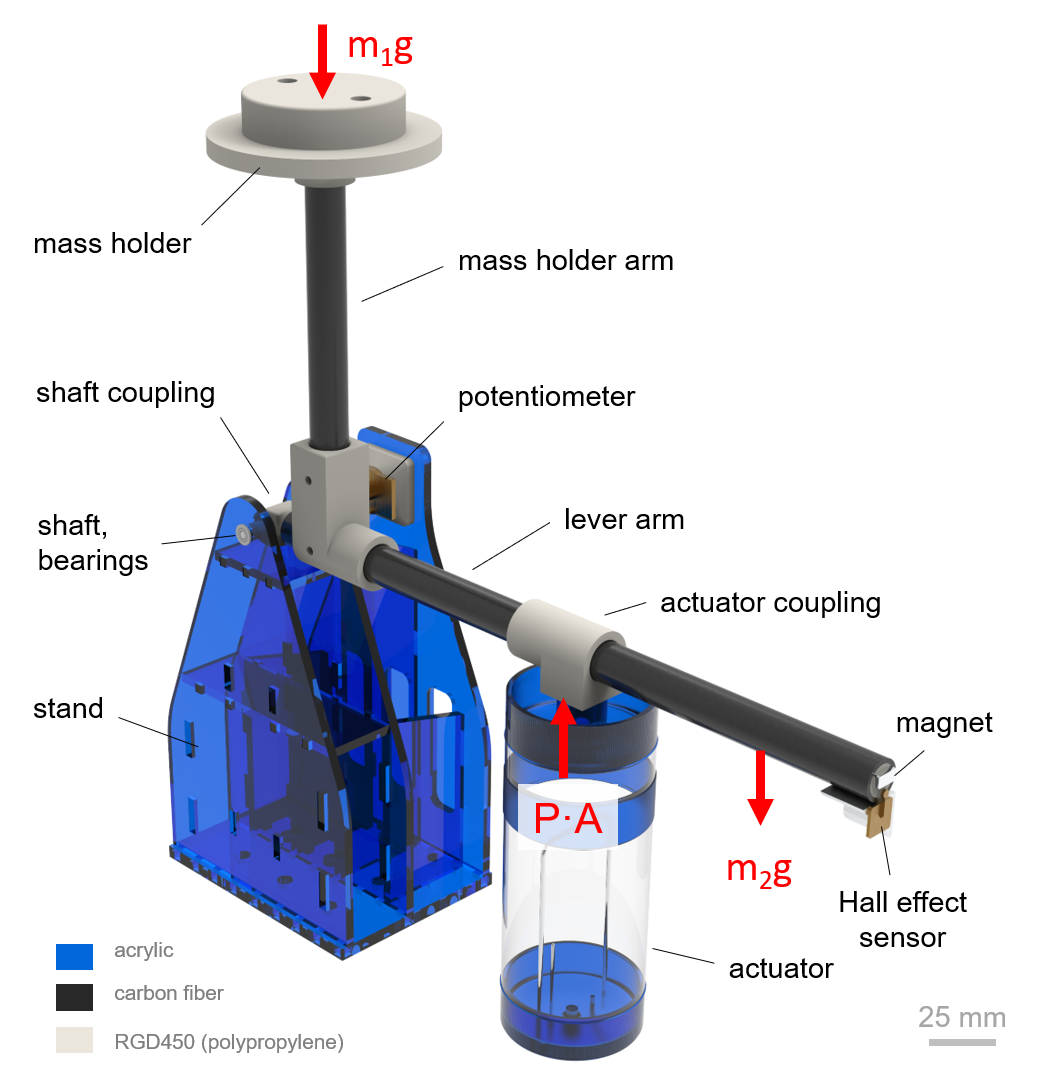}
  \caption{Actuator testing setup.}
  \label{fig:Test_Setup}
  \end{center}
 \end{figure}

A custom actuator testing apparatus was designed and built (Fig. \ref{fig:Test_Setup}) to produce approximate constant-load (Fig. \ref{fig:isoBarPV}) and decreasing-load (Otto-like) (Fig. \ref{fig:ottoPV}) cycles. An acrylic stand was laser-machined and supports a steel rotation shaft housed by ball bearings. 3D-printed components (\textit{Stratasys Objet30}, RGD450) couple this shaft (1) to a carbon fiber lever arm that translates the vertical motion of the actuator into rotational motion of the shaft, and (2) at its end to a potentiometer that provides continuous angle measurement. Here, change in lever angle serves as an easy-to-measure proxy for actuator volume, because the actuator nominally only expands in one dimension (height); thus, we present pressure-angle plots herein as proxies for pressure-volume plots. 

\begin{figure}
  \begin{center}
  \includegraphics[width=\columnwidth]{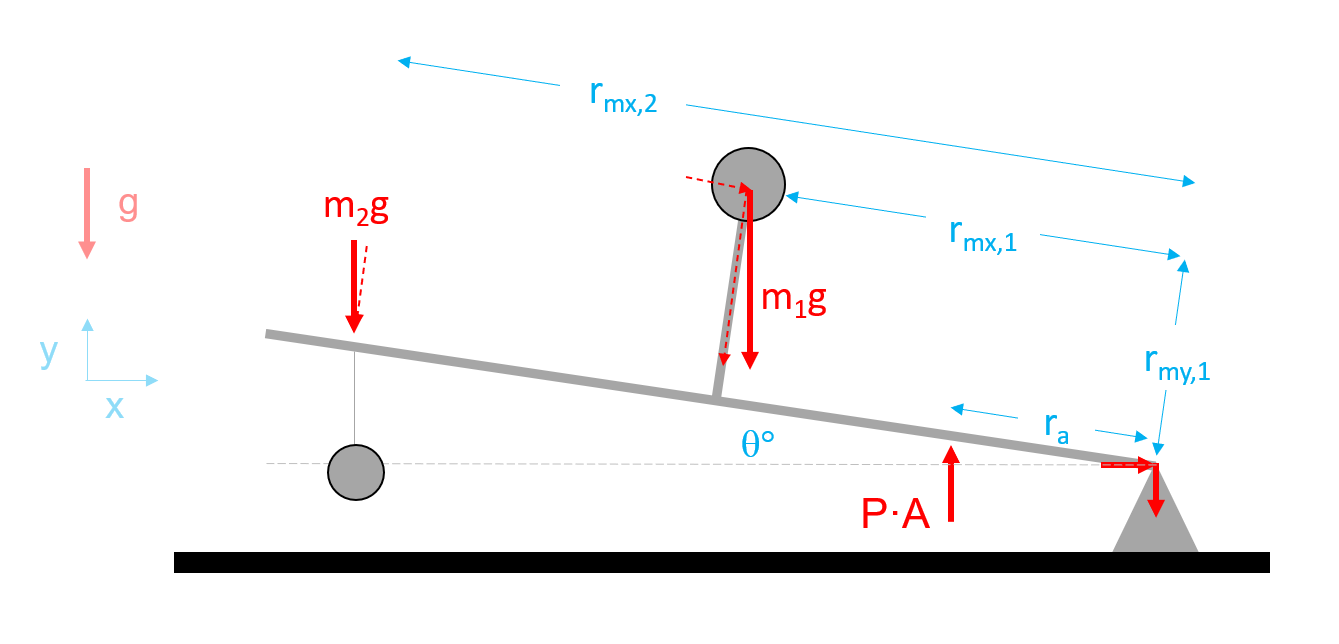}
  \caption{A free-body-diagram of the testing setup.}
  \label{fig:free}
  \end{center}
 \end{figure}

The load curve for both constant-load and decreasing-load cycles can be calculated by solving for actuator pressure $P$ via a free-body-diagram of the testing setup (Fig. \ref{fig:free}). For a constant-load cycle, no mass is attached perpendicular to the lever arm (i.e., $m_1 = 0$, Fig. \ref{fig:free}). The constant-load expansion (i.e., 2-3, Fig. \ref{fig:isoBarPV}) requires a mass $m_2$ hung from the cantilever at $r_{mx,2}$ that is larger than the mass $m_2$ required for the restoring load (i.e., 4-1, Fig. \ref{fig:isoBarPV}). I.e., the following condition must be met for positive work in the constant-load cycle:

\[(P_{2-3})_{Iso} = \frac{m_{2,2-3}g\cos\theta r_{mx,2}}{Ar_a} >\]
\[(P_{4-1})_{Iso} = \frac{m_{2,4-1}g\cos\theta r_{mx,2}}{Ar_a}\]

Where \(m_{2,2-3} > m_{2,4-1}\). In practice, this is achieved by adding/removing mass at different time points in the constant-load test, while all other variables remain constant. 

For the decreasing-load cycle, a mass was fixed perpendicular to the rotating lever arm (see ``mass holder", Fig. \ref{fig:Test_Setup}). The Otto-like restoring load (i.e., 1-2, Fig. \ref{fig:ottoPV}) is achieved by solving for another expression of actuator pressure $P$ from the free-body-diagram in Fig. \ref{fig:free} and calculating the necessary mass $m_1$ and its horizontal $r_{mx,1}$ and vertical $r_{my,1}$ distance from the axle. The Otto-like expansion load (i.e., 3-4 in Fig. \ref{fig:ottoPV}), is achieved by calculating the requisite mass $m_2$ to be added/removed. I.e., the following condition must be met for positive work in the Otto cycle:

\[(P_{3-4})_{Otto} =  \frac{m_1g\cos\theta r_{mx,1} - m_1g\sin\theta r_{my,1} + m_2g\cos\theta r_{mx,2}}{Ar_a} >\]
\[(P_{1-2})_{Otto} =  \frac{m_1g\cos\theta r_{mx,1} - m_1g\sin\theta r_{my,1}}{Ar_a}\]

\subsection{Actuator Design}

\begin{figure}
  \begin{center}
  \includegraphics[width=\columnwidth]{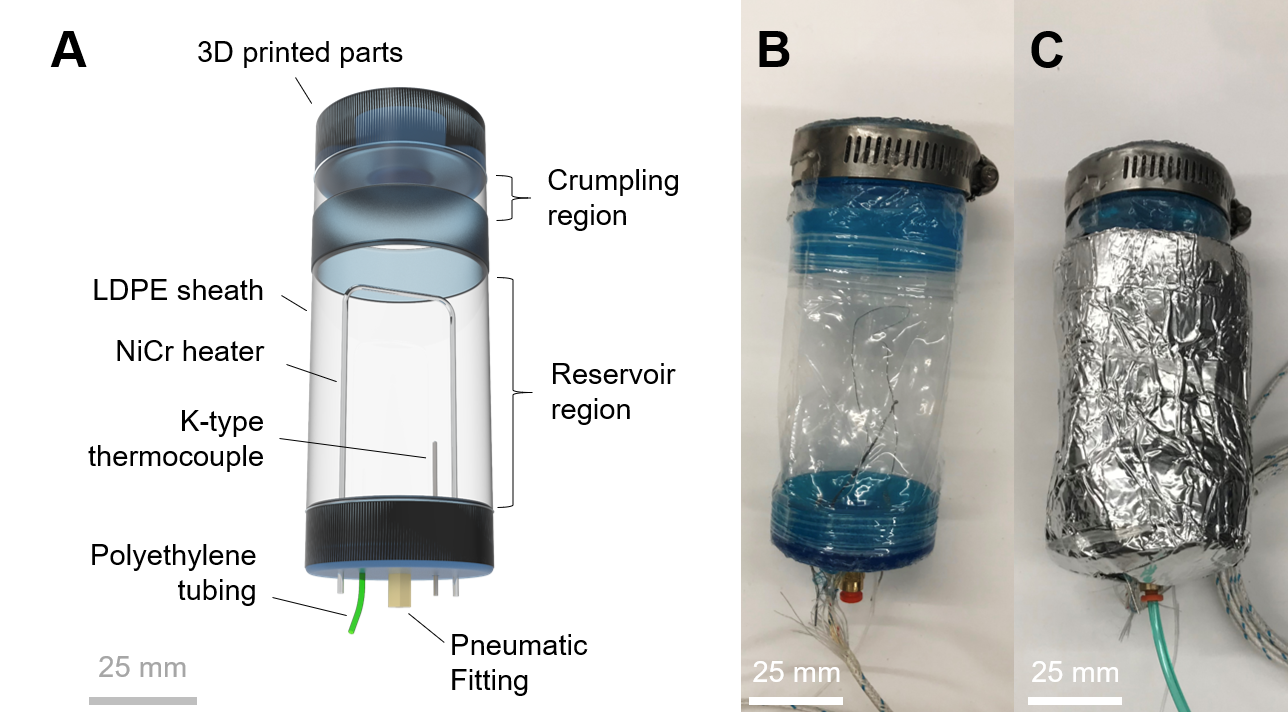}
  \caption{\textbf{A}, Labeled 3D rendering of the actuator. \textbf{B}, Depiction of real actuator. \textbf{C}, Actuator with insulation.}
  \label{fig:actuator}
  \end{center}
\end{figure}

The soft actuator's body is comprised of \SI{.0254}{\milli\meter} thick, \SI{48}{\milli\meter} inner-diameter low-density polyethylene (LDPE) lay-flat tubing (Fig. \ref{fig:actuator}A). To provide structure and mounting points for testing, three acrylic cylindrical structures were 3D-printed (\textit{FormLabs Form 2}, Tough Resin) and connected to the inside of the pouch via press fit and/or double-sided tape (for top and bottom structures). On the outside, fiberglass tape was wrapped around the structures to secure the connection between the pouch and the structures. The top of the pouch was secured using a hose clamp and leak-prone regions were sealed with epoxy. Routed through the bottom of the actuator are: (1) a nickel-chromium wire heating element, (2) a K-type thermocouple, (3) a brass pneumatic fitting that connects a \SI{3.2}{\milli\meter} diameter tubing from the pressure sensor to the actuator, and (4) \SI{0.9}{\milli\meter} diameter polyethylene (PE) tube terminated by a valve which can be opened and connected to a syringe for inflation/deflation. The routing holes were sealed with epoxy. Lastly, two layers of \SI{3}{\milli\meter} neoprene foam and one layer of aluminum foil tape were wrapped around the actuator's reservoir region to minimize heat losses (Fig. \ref{fig:actuator}C). Once the actuator was complete, we measured the force it takes to deform the walls of the actuator. We do this by keeping the ports of actuator open and slowly compressing and expanding the actuator. A force of \SI{1.3}{\newton} was measured. This is small relative to the loads seen during experimentation.

\subsection{Load Design}

\begin{figure}
    \begin{center}
  \includegraphics[width=\columnwidth]{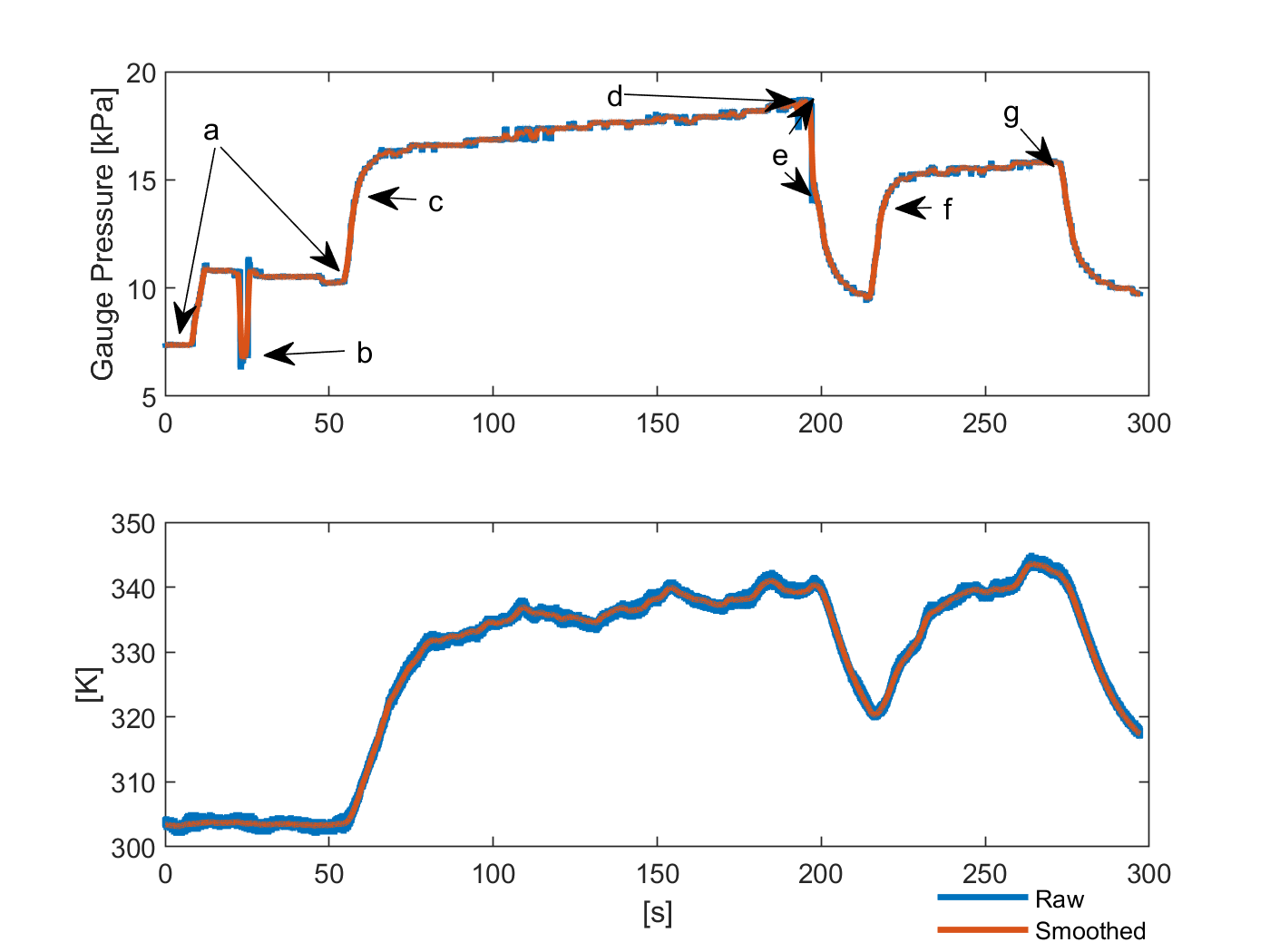}
  \caption{Actuator characterization tests.}
  \label{fig:limTest}
  \end{center}
\end{figure}

The force output of the actuator must be characterized to design the appropriate load profiles for both the constant-load and decreasing-load cycles. This is achieved via an actuator characterization test that offers several key insights (Fig. \ref{fig:limTest}). Prior to this test: (1) the air supply valve is opened and air is dispensed into the actuator until a pressure of \SI{6.9}{\kilo\pascal} is reached, (2) the valve is then closed, (3) the stroke of the actuator is limited to \SI{1.4}{\degree} above horizontal, and (4) the heater was supplied with \SI{3.57}{\watt} via a direct current (DC) voltage supply.

\begin{enumerate}
    \item Isothermal expansion ratio: System pressure is measured at zero stroke and again at maximum stroke after a slow expansion, and estimated via an ideal gas relation:
    \[\frac{P_1}{P_2}=\frac{V_2}{V_1}\]
    Label \textit{a} in Fig. \ref{fig:limTest} illustrates $P_1$ and $P_2$. We estimate an expansion ratio of 1.027.
    \item Isentropic compression estimate: At label \textit{b} the actuator was rapidly compressed. A pressure rise of about \SI{5.0}{\kilo\pascal} is estimated.
    \item Otto cycle maximum pressure estimate: The actuator is locked in the minimum stroke state and heated. At label \textit{d}, a maximum pressure of \SI{18.0}{\kilo\pascal} was estimated for $P_3$. However, for our tests, we will target a maximum pressure at label \textit{c} of \SI{14.0}{\kilo\pascal}, as achieving the pressure at \textit{d} is less efficient due to large heat losses.
    \item Isentropic expansion estimate: Process 3-4 on the Otto cycle is estimated by rapidly expanding the actuator to maximum stroke immediately after turning off the heater. Label \textit{e} spans a range of about \SI{4.5}{\kilo\pascal}.
    \item Constant-load cycle maximum pressure: the heater is left on at maximum stroke. Label \textit{g} suggests a maximum pressure of about \SI{15.0}{\kilo\pascal}. However, again, we target a maximum pressure at label \textit{f} of about \SI{13}{\kilo\pascal} because it is in a lower heat loss regime.
    \item Constant-load cycle minimum pressure: To ensure that the cycle can return to the compressed state at ambient temperatures, the restoring load target is about \SI{11.0}{\kilo\pascal} (label \textit{a}, right arrow).
\end{enumerate}



\subsection{Data Collection}
The pressure inside of the actuator was measured using a \textit{Dwyer} 628-10-GH-P1-E1-S1 pressure transmitter. Forces were measured with a \textit{Mark-10} M3-5 force gauge. Air temperature inside of the actuator was measured using a K-type thermocouple. Lever angle was measured using a \SI{50}{\kilo\ohm} rotary potentiometer. A permanent magnet was adhered to the end of the lever, triggering a proximity (i.e., Hall effect) sensor. This trigger is processed by an Arduino Uno, which then switches a voltage relay and sends current to the actuator's Joule heater. Voltage signals from the pressure sensor, thermocouple, potentiometer, and switch relay were all sampled at a channel rate of \SI{50}{\hertz} using a \textit{DATAQ DI-2008}. 

\subsection{Cycle Testing}
For cycle testing, the actuator was manually loaded appropriately. A DC power supply provided constant power to the heater, with voltage input set to \SI{3.92}{\volt} and current varying between \SI{.90}{\ampere} and \SI{.91}{\ampere}. The heating was controlled via Arduino and the temperature, pressure, angle, and power input data recorded. Only data recorded after several cycles were analyzed to minimize any transient effects upon initial warming.  

\section{Experimental Validation}

In this section, we present the data from our cycle tests.

\subsection{Constant Load Test}
\begin{figure}
  \begin{center}
  \includegraphics[width=\columnwidth]{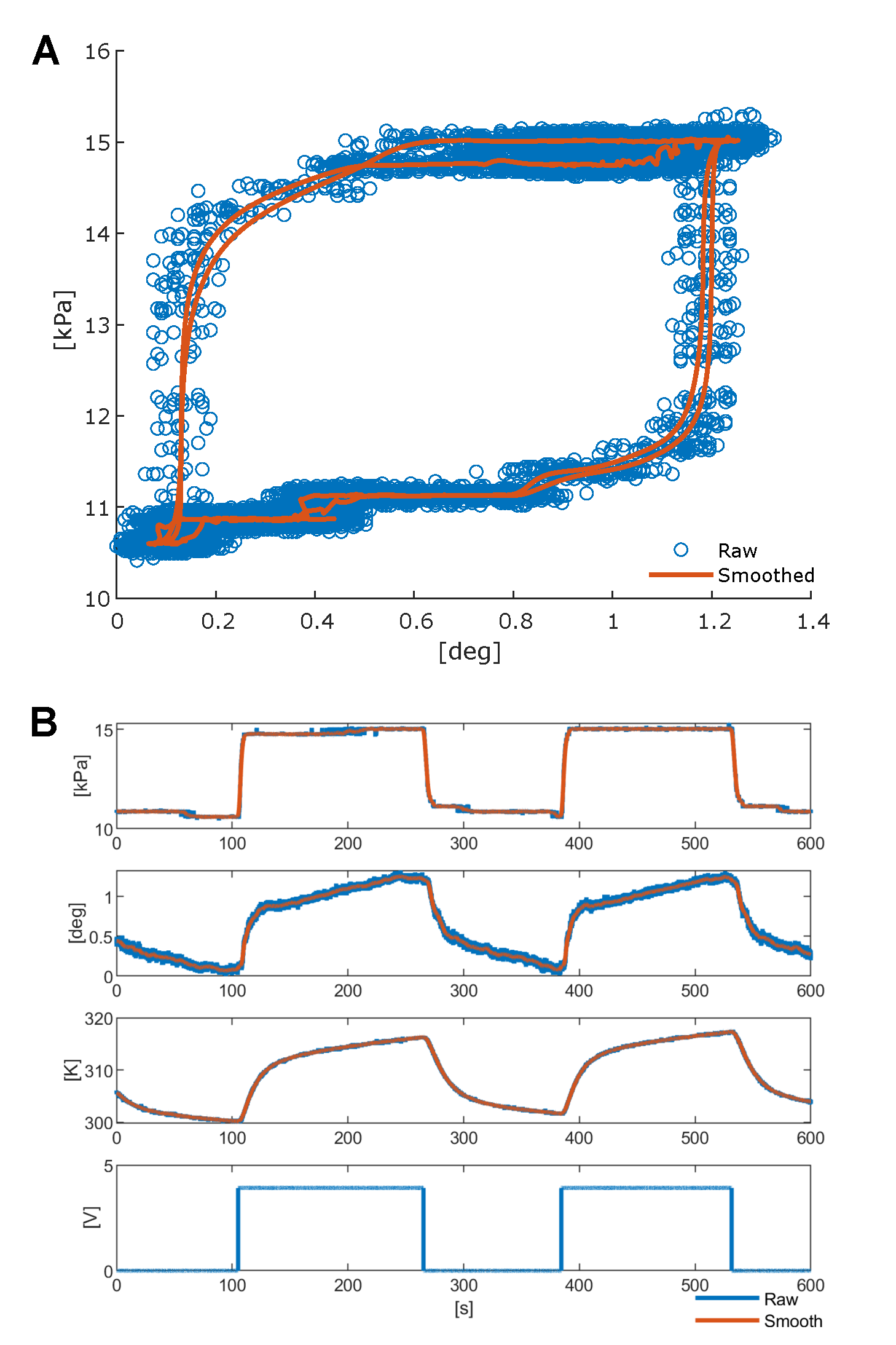}
  \caption{Data from two representative constant-load cycles.  \textbf{A}, Pressure-angle data. \textbf{B}, Time-dependent angle, pressure, temperature and heater state data from the two cycles.}
  \label{fig:isoBarPV}
  \end{center}
 \end{figure}
 
 Fig. \ref{fig:isoBarPV}A shows the pressure-angle (PA) data for two exemplar constant load cycles and Fig. \ref{fig:isoBarPV}B shows the angle, pressure, temperature, and heater state data as a function of time for those cycles. Readings of \SI{4}{\volt} and  \SI{0}{\volt} indicate that the heater is on and off, respectively. A restoring load (i.e., 4-1 in Fig. \ref{fig:isoBarPV}) of  \SI{1.361}{\kilo\gram} and a expansion load (i.e., 2-3 in Fig. \ref{fig:isoBarPV}) of \SI{1.643}{\kilo\gram} were applied \SI{200}{\milli\meter} from the lever axle.
 
 The maximum stroke of the lever, determined from smoothed data, was \SI{1.2}{\degree}, which translates to a mass displacement of \SI{4.2}{\milli\meter}. This corresponds to net work of \SI{0.012}{\joule}. During the cycle, the gauge pressure inside the actuator varied between \SI{10.6}{\kilo\pascal} and \SI{15.0}{\kilo\pascal}.  The average heating duration was \SI{153.6}{\second}, resulting in a high estimate heat input of \SI{547.9}{\joule}. Temperatures varied between \SI{300}{\kelvin} and \SI{317}{\kelvin}.  The calculated efficiency of the cycle is 0.0021\%.
 
 \subsection{Otto-like Cycle Test}
 
 \begin{figure}
    \begin{center}
  \includegraphics[width=\columnwidth]{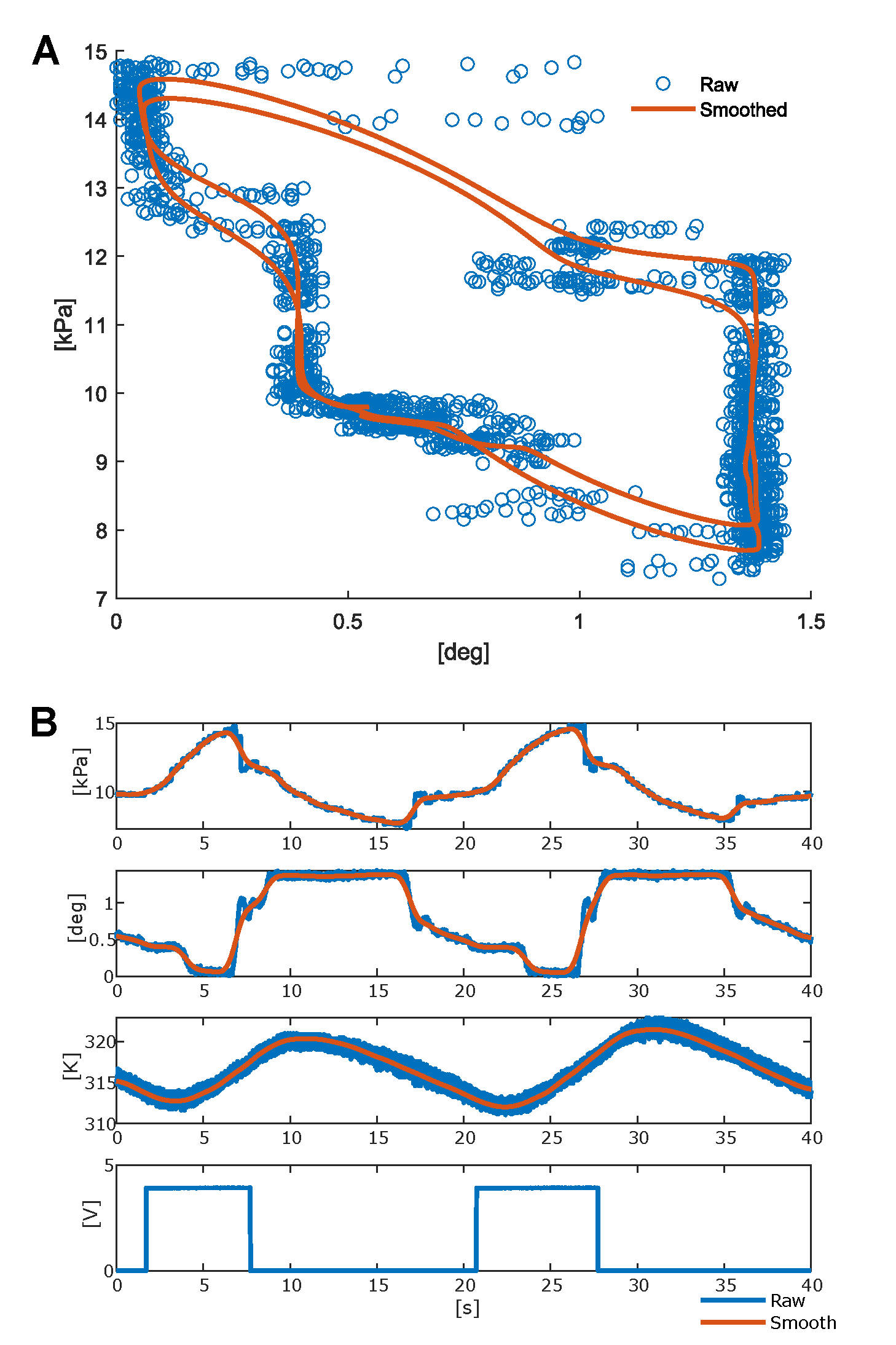}
  \caption{Data from two representative decreasing-load Otto-like cycles. \textbf{A}, Pressure-angle data. \textbf{B}, Time-dependent pressure, angle, temperature, and heater state data from the two cycles.}
  \label{fig:ottoPV}
  \end{center}
\end{figure}

 Fig. \ref{fig:ottoPV}A illustrates PA data for two representative Otto-like cycles and Fig. \ref{fig:ottoPV}B shows the angle, pressure, temperature, and heater state data as a function of time for those cycles. Restoring load $m_1$ = \SI{13.6}{\kilo\gram} was applied at approximately $r_{mx,1}$ = \SI{28}{\milli\meter} and $r_{my,1}$ = \SI{210}{\milli\meter} from the axle. During expansion, a load $m_2$ = \SI{1.00}{\kilo\gram} was suspended at $r_{mx,2}$ = \SI{28}{\milli\meter} from the axle.
 
 The maximum stroke in this case is \SI{1.3}{\degree}, which translates to a mass displacement of \SI{0.64}{\milli\meter}; resulting in a net work of \SI{0.0062}{\joule}. The gauge pressure inside the actuator varied between \SI{7.4}{\kilo\pascal} and \SI{14.6}{\kilo\pascal} across all tests.  The average heating duration was \SI{5.5}{\second} resulting in an average high estimate heat input of  \SI{19.6}{\joule}. Across all tests, temperatures varied between \SI{310}{\kelvin} and \SI{321}{\kelvin}. The calculated efficiency of the cycle is 0.032\%.

\section{Discussion}

In this work, we applied thermodynamic theory to analyze the efficiency of air-based soft heat engines. One key outcome of this theory is the prediction that an Otto-like engine designed with a decreasing-load cycle should outperform a constant-load cycle, due to larger amount of non-isothermal heat transfer in the latter case. Through simple experiments, we tested this prediction and validated the model, measuring a factor of 11.3 improvement when the Otto-like cycle was used. 

In reviewing the particular details of the experiments presented above, we gain additional insight into the cycle design and actuator performance. In practice, we found that heat losses were a major source of inefficiency for both cycles. In fact, the heat losses were so significant that the working stroke of the actuator was limited to half of its full range. From our estimated expansion ratio, we expected the efficiency of the decreasing-load Otto-like cycle to be on the order of 1.0\%, which is nearly two orders of magnitude more than the measured efficiency of 0.032\%. This suggests better insulation is needed and that the isentropic processes need to happen faster, but also indicate the great potential for improved performance of such air-based soft actuators in future designs. Moreover, this observation may partially explain the relatively high efficiency of the combustion-based soft jumping robots. Beyond the higher temperatures they achieve, their expansions are explosive in nature and thus occur more adiabatically. 

We also note that there was an actuator repair between the Otto-like and constant-load tests.
This repair process led to minor variability in the stroke of the actuator, and a minor shift in the thermocouple position that slightly shifted reported temperatures.
In comparison to the data from Fig. \ref{fig:limTest} we believe that the data shown in Fig. \ref{fig:isoBarPV} underreport the temperatures that would have been measured in the constant-load tests were it not for the temperature sensor re-positioning. 

While our results lay the critical groundwork for the analysis and design of air-based soft heat engines, future investigations are needed, in particular to: study the heat transfer process in soft heat engines, investigate load design in phase-change soft robot systems, and devise improved adiabatic processes. Investigating adiabatic processes could be particularly fruitful, because of the growing interest in snap-through mechanisms within the soft robotics community. Such mechanisms have very fast actuation \cite{stadlbauer2021body, gorissen2020inflatable} and have to some extent already been studied for refrigeration applications \cite{greibich2021elastocaloric}.

\section{Conclusion}

Here, we explored the theoretical and experimental use of air as a working fluid for soft heat engines. Theoretically, we found that the constant-load cycle is inefficient for air-based systems due to the high degree of non-isothermal heat transfer. To limit the amount of non-isothermal transfer, we propose using an Otto-like cycle instead, which replaces the constant-load processes with isentropic ones. Theoretically, this is more efficient for a given temperature and expansion ratio. Critically, a decreasing-load profile is required to approximate an isentropic expansion. Experimentally, we found that the air-based soft heat engine running an Otto-like cycle is 11.3 times more efficient than one running a constant-load cycle. 





\bibliographystyle{IEEEtran}
\bibliography{IEEEabrv,Bibliography}

\end{document}